\documentclass[sigconf]{acmart}
\pdfoutput=1
\copyrightyear{2022}
\acmYear{2022}
\setcopyright{acmcopyright}
\acmConference[MM '22] {Proceedings of the 30th ACM International Conference on Multimedia }{October 10--14, 2022}{Lisboa, Portugal.}
\acmBooktitle{Proceedings of the 30th ACM International Conference on Multimedia (MM '22), October 10--14, 2022, Lisboa, Portugal}
\acmPrice{15.00}
\acmISBN{978-1-4503-9203-7/22/10}
\acmDOI{10.1145/3503161.3548109}

\acmSubmissionID{1517}

\usepackage{multirow}
\usepackage{bm}
\usepackage{amsmath}
\usepackage{algorithm}
\usepackage{algorithmicx}
\usepackage{booktabs}
\usepackage{bbding}
\usepackage{xspace}
\usepackage{algpseudocode}

\begin{document}

\title{Design What You Desire: Icon Generation from Orthogonal Application and Theme Labels}

\author{Yinpeng Chen}
\email{yinpengchen@hust.edu.cn}
\authornote{These authors contributed equally to this work.}
\affiliation{
  \institution{School of AIA, Huazhong University of Science and Technology}
  \country{}
}

\author{Zhiyu Pan}
\email{zhiyupan@hust.edu.cn}
\authornotemark[1]
\affiliation{
   \institution{School of AIA, Huazhong University of Science and Technology }
  \country{}
}

\author{Min Shi}
\email{min_shi@hust.edu.cn}
\affiliation{
   \institution{School of AIA, Huazhong University of Science and Technology}
  \country{}
}

\author{Hao Lu}
\email{hlu@hust.edu.cn}
\authornote{Corresponding author.}
\affiliation{
   \institution{School of AIA, Huazhong University of Science and Technology}
  \country{}
}

\author{Zhiguo Cao}
\email{zgcao@hust.edu.cn}
\affiliation{
   \institution{School of AIA, Huazhong University of Science and Technology}
  \country{}
}

\author{Weicai Zhong}
\email{zhongweicai@huawei.com}
\affiliation{
  \institution{CBG Search \& Maps BU, Huawei Inc.}
  \country{}
}

\renewcommand{\shortauthors}{Yinpeng Chen et al.}

\def\dataset{{AppIcon}\xspace}

\begin{teaserfigure}
\vspace{-7pt}
    \begin{minipage}[b]{.28\linewidth}
      \centering
      \centerline{\includegraphics[width=\linewidth]{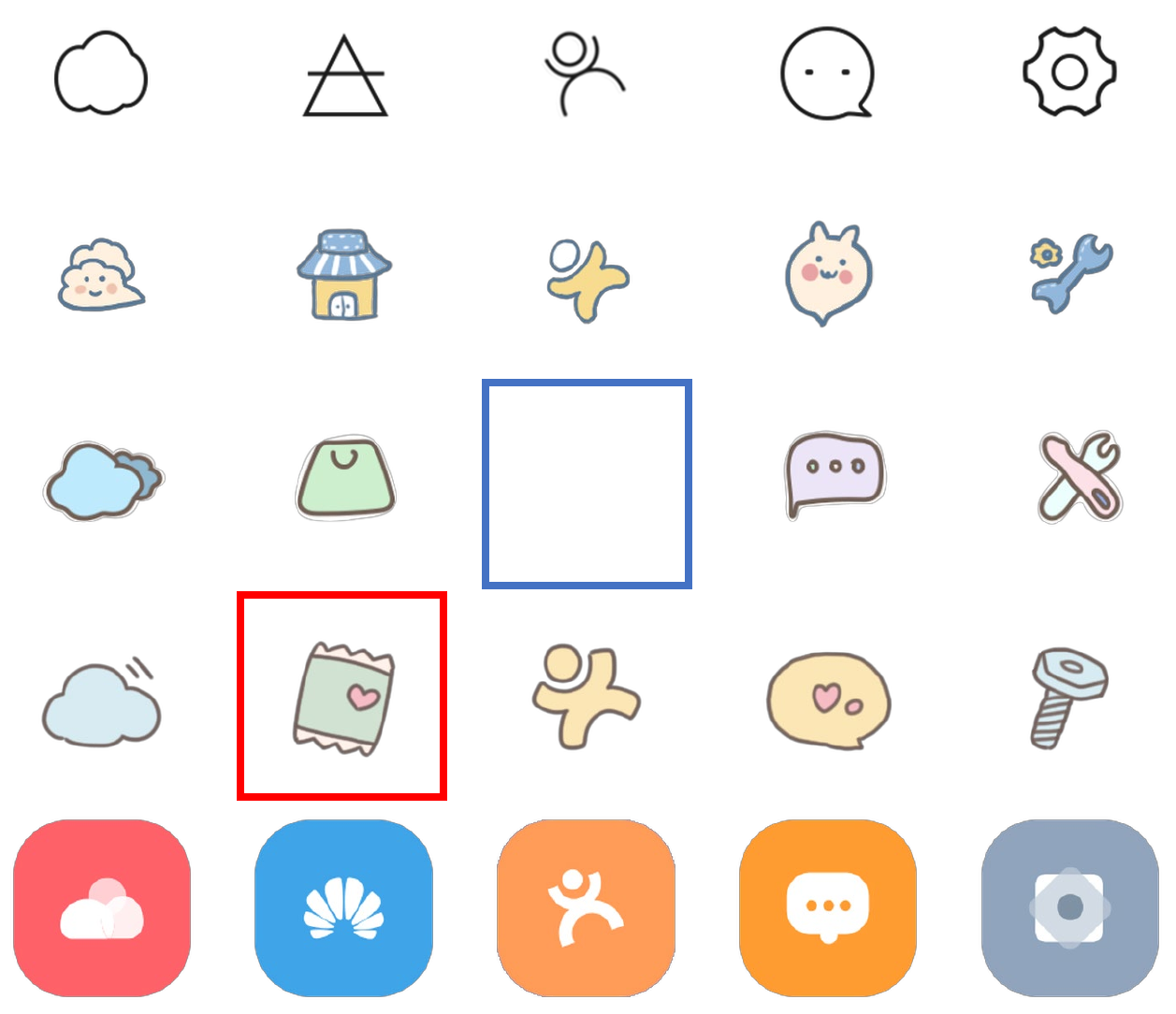}}
      \centerline{\textbf{(a) \dataset Dataset\quad}}
    \end{minipage}
    \hfill
    \begin{minipage}[b]{.28\linewidth}
      \centering
      \centerline{\includegraphics[width=\linewidth]{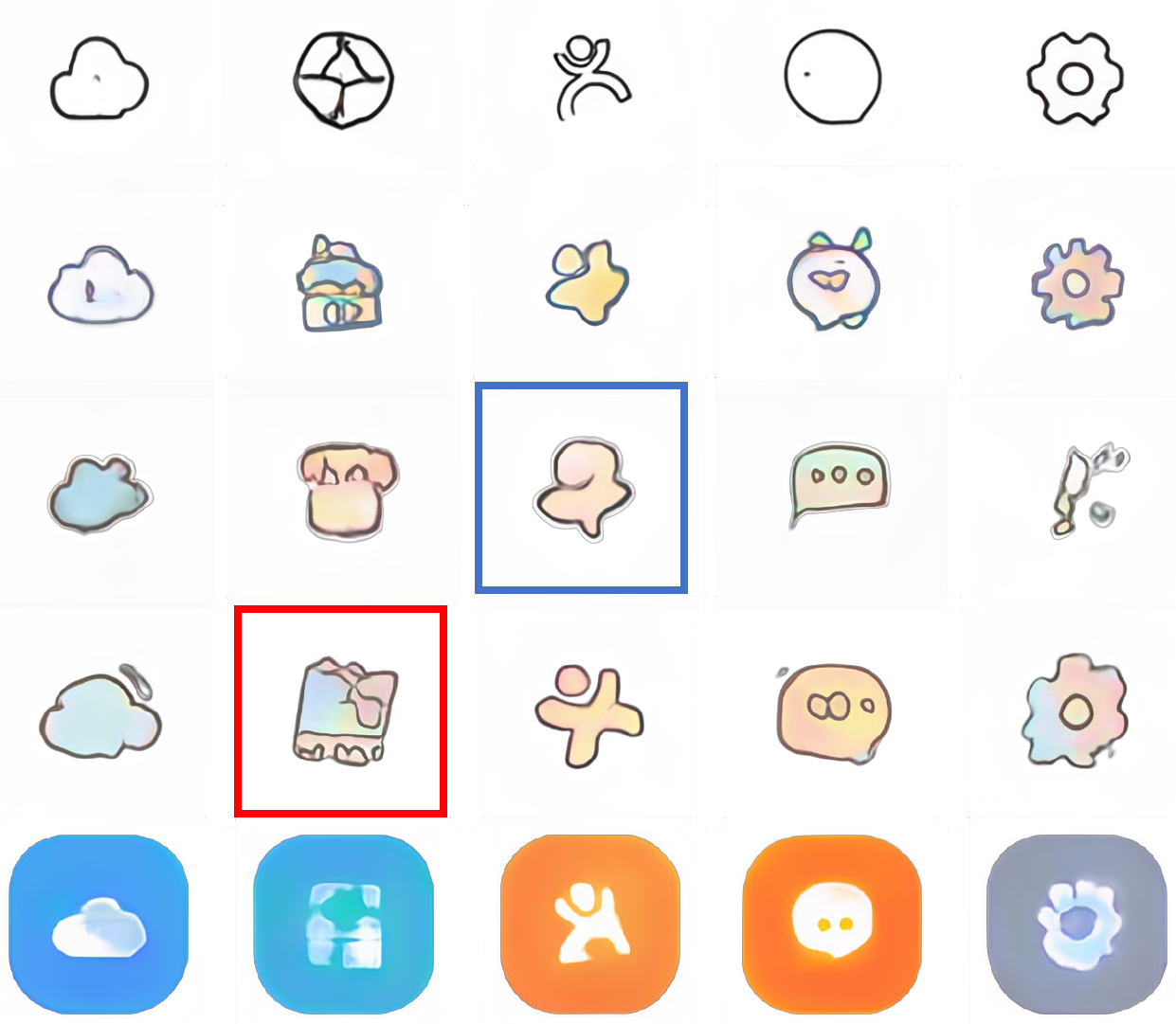}}
      \centerline{\textbf{(b) StyleGAN2}}
    \end{minipage}
    \hfill
    \begin{minipage}[b]{.28\linewidth}
      \centering
      \centerline{\includegraphics[width=\linewidth]{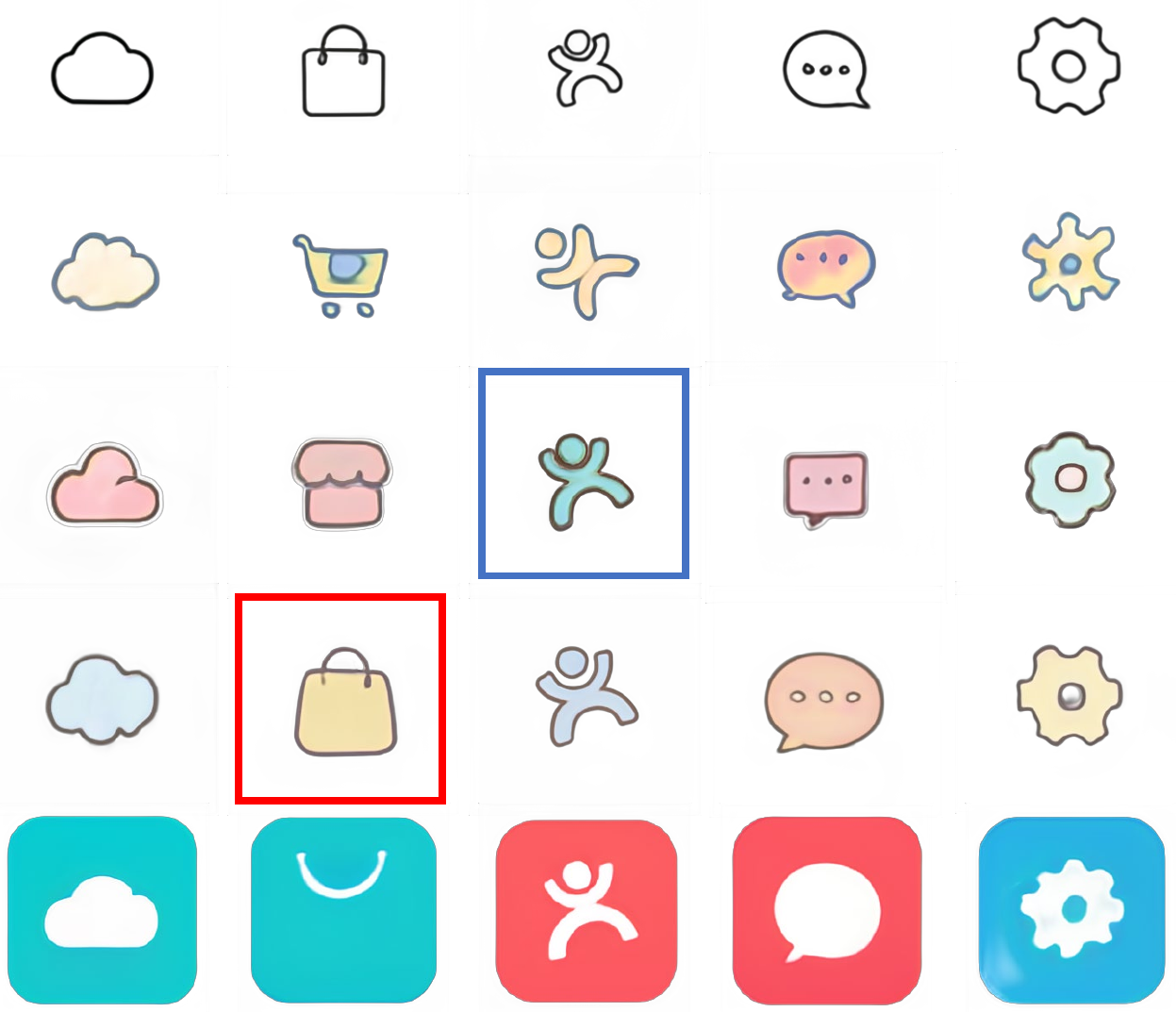}}
      \centerline{\textbf{\quad(c) IconGAN (Ours)}}
    \end{minipage}
    \vspace{-5pt}
  \caption{Designing icons that you desire. (a) Icons from the proposed \dataset dataset. The icons in the same column and row share the same app and theme label, respectively. The blanks denote there is no such an icon for the given app-theme combination. (b) Icons generated by StyleGAN2 tend to only learn the superficial representation of icons in the dataset. (c) Icons generated by our IconGAN designs new icons conditioned on the combination of learned app and theme representations. The blue and red boxes highlight two different cases where StyleGAN2 generates icons with unreasonable contents.
  }
  \label{fig:teaser}
  \vspace{5pt}
\end{teaserfigure}

\begin{abstract}
Generative adversarial networks\,(GANs) have been trained to be professional artists able to create stunning artworks such as face generation and image style transfer. In this paper, we focus on a realistic business scenario: automated generation of customizable icons given desired mobile applications and theme styles. We first introduce a theme-application icon dataset, termed \dataset, where each icon has two orthogonal theme and app labels. By investigating a strong baseline StyleGAN2, we observe mode collapse caused by the entanglement of the orthogonal labels. To solve this challenge, we propose IconGAN composed of a conditional generator and dual discriminators with orthogonal augmentations, and a contrastive feature disentanglement strategy is further designed to regularize the feature space of the two discriminators. Compared with other approaches, IconGAN indicates a superior advantage on the \dataset benchmark. Further analysis also justifies the effectiveness of disentangling app and theme representations. Our project will be released at: \href{https://github.com/architect-road/IconGAN}{https://github.com/architect-road/IconGAN}.
\end{abstract}



\maketitle

\section{Introduction}
People have been playing with various smartphone applications in their everyday life. Suddenly one day one may be tried of looking at the same app icon and decides to make some changes such as switching to a personalized theme. Once the theme is set, one may disappointingly find that some icons remain unchanged as they are not designed in the theme. The unchanged icons seem like square pegs in round holes, and somehow the icon designer would be made a scapegoat. The designer, however, should not be blamed. Icon designing is expensive, because a well-designed icon should not only conform to the style of the theme but also inform the function of the application. Nowadays hundreds of new apps are produced every day; it is infeasible to achieve icon designing and completion over a large number of apps with only manual efforts. The need for large-scale, theme-consistent icon designing therefore poses a new challenge to our community: can high-quality icon designing be automated by machine?

Having achieved tremendous success in creating various artworks~\cite{isola2017image, cheng2020rifegan, wen2021zigan}, generative adversarial networks \,(GANs)~\cite{goodfellow2014generative} can be a preferable choice for designing icons. In this work, we investigate the potential of GANs on this novel icon designing task, \textit{i.e.}, designing style-consistent and app-informative icons given target themes and applications. This task can also be extended to other scenarios where user interfaces or icons need to be designed.

However, we find this task confronted with two obstacles: 1) there is no such an icon dataset annotated with both apps and themes; 2) the existing cutting-edge GAN model suffers from mode collapse due to the entanglement of theme style and app content, as shown in Fig.~\ref{fig:teaser}\,(b). To address both issues, we first elaborate a high-quality and well-organized icon dataset named \dataset, where each icon is annotated with two orthogonal labels: app and theme. Some samples are exemplified in Fig.~\ref{fig:teaser}\,(a). Here the orthogonal labels mean the implication and distribution of two labels are completely irrelevant. And there is only one icon given two certain app and theme labels in our \dataset dataset which renders great challenges for a vanilla conditional GAN.

To investigate whether existing GANs can be directly adopted, we test a strong baseline based on conditional StyleGAN2~\cite{karras2020analyzing}, where conditions are conveyed by concatenating the theme and app labels. As shown in Fig.~\ref{fig:teaser}\,(b), in most cases, the model simply learns to reproduce the original icons in the dataset and cannot create new app-informative content. If an app-theme label does not hold any icon in the dataset or corresponds with an outlier which is hard to remember, the generation result shows obvious artifacts. This indicates that the model only learns the superficial representation related to the combined labels rather than the intrinsically orthogonal app and theme representations for each icon. In another word, the model regards the combination of theme and app labels as a joint condition, and fails to disentangle them as two orthogonal labels. By observing the artifacts caused by the entanglement of app and theme labels in the baseline, we hypothesize the crux of our theme-app icon generation task is \textit{how to disentangle the representations for apps and themes}.

To this end, we present a novel icon generation model named IconGAN. Its core idea is to disentangle the learning of app and theme representations. The disentanglement is achieved via a dual-discriminator architecture and a novel contrastive feature disentanglement loss. In particular, two different discriminators are employed to distinguish whether the theme and app of the generated icons are consistent with the input labels. We also develop a heuristic augmentation strategy to enhance the data diversity for discriminators to understand contents and styles. In this way, the app-related contents and theme styles can be independently encoded. To decouple the encoded app and theme features, we further devise a contrastive feature disentanglement loss where icon app features with the same app labels are encouraged to be close while their theme features are required to be separated. On the contrary, icons under the same theme are constrained to possess similar theme features but differing app features. 

Extensive experiments on our dataset show that, the proposed IconGAN can create various app-informative contents in harmony with the theme styles, as shown in Fig.~\ref{fig:teaser}\,(c). Ablation studies and the visualizations also demonstrate the effectiveness of our disentanglement strategies for app contents and theme styles. Furthermore, we visualize the feature distribution of the encoded app and theme features. The results show our disentanglement strategy reaches its goal: training a GAN model that is truly aware of the app-related contents and theme styles.

Our contributions are three-fold:

$\bullet$ We introduce a novel task -- icon generation conditioned on both app and theme labels.

$\bullet$ We collect a high-quality and well-organized icon dataset, including $23,164$ icons from $476$ themes and $52$ mobile applications.

$\bullet$ We propose a novel icon generation model with dual discriminators and contrastive feature disentangling, termed IconGAN.

\section{Related Work}
\noindent\textbf{Conditional Generative Adversarial Networks.}
Compared with the vanilla GAN, conditional GAN\,(cGAN) \cite{mirza2014conditional} aims to generate more controllable contents given some conditions, \textit{e.g.}, class conditional image generation~\cite{odena2017conditional, miyato2018cgans, kang2020contragan, kang2021rebooting}, text-based image generation~\cite{cheng2020rifegan,li2019controllable, zhang2021cross} and image-to-image translation~\cite{park2020contrastive,isola2017image,zhu2017unpaired}. In early stage, AC-GAN~\cite{odena2017conditional} proposes to use an auxiliary classifier to discriminate whether the generated images accord with input conditions. However, employing such an auxiliary classifier may unwittingly cause degeneration to trivial solutions. Hence, Miyato~\textit{et al.} propose projection discriminator~\cite{miyato2018cgans} that directly constrain the condition and the discriminator representation based on a novel projection way. More recently, to help model distinguish different conditions in feature space, ContraGAN~\cite{kang2020contragan} and ReACGAN~\cite{kang2021rebooting} adapts contrastive loss~\cite{hadsell2006dimensionality} into the training of GANs. In icon generation, our condition is the orthogonal labels w.r.t.\ app and theme. The proposed IconGAN hence inherit the vein of cGAN. However, we find simply adopting existing cGAN models or ideas struggles from the entanglement of two labels, which yields inferior results.

\vspace{5pt}
\noindent\textbf{Icon Generation.}
Unlike real-world images or paintings, icons or logos are built from neat, clear lines and color blocks, and display more abstract patterns. Compared with natural images that have smooth distributions, the distribution for icon/logo data is more discrete, which poses great challenges for GAN-based icon generation. Recently, some approaches~\cite{sage2018logo, mino2018logan, oeldorf2019loganv2, mao2020computerized, dong2022generative} attempt to employ GANs for icon/logo generation. Sage \textit{et al.}~\cite{sage2018logo} uses synthetic labels obtained from heuristic clustering to disentangle and stabilize the training process. LoGANs~\cite{mino2018logan, oeldorf2019loganv2} generate icons based on main color and synthetic class conditions. Yang~\textit{et al.}~\cite{yang2021icon} directly employs the StyleGAN~\cite{karras2019style} with the self-attention mechanism~\cite{zhang2019self} to generate icons conditioned on $8$ application categories. Different from the GAN-based approaches, DeepSVG~\cite{carlier2020deepsvg} designs a hierarchical transformer-based architecture to generate icons in a scalable vector graphics format. However, none of previous literature attempts to generate icons conditioned on both app and theme labels, which mostly aligns with the realistic business scenario. Here we propose a well-organized icon dataset with abundant app and theme labels. With the help of the dataset, we generate high-quality icons successfully under full control of app and theme.

\vspace{5pt}
\noindent\textbf{Disentanglement in GANs.}
In open literature, three typical types of disentanglement are mainly studied: input space, content\&style, and attribute disentangling.
For the disentanglement of input latent space, InfoGAN~\cite{chen2016infogan} maximizes the mutual information between part of latent codes and generation results. StyleGAN~\cite{karras2019style} proposes mapping network to project random latent codes to a more linear sub-space. And for the content\&style disentangling, which originated from style transfer~\cite{gatys2016image, huang2017arbitrary, johnson2016perceptual}, Mathieu~\textit{et al.}~\cite{mathieu2016disentangling} and Lee \textit{et al.}~\cite{lee2018diverse} disentangle the sub-space of content and style by adding an adversarial constraint to the encoders. Kotovenko \textit{et al.}~\cite{kotovenko2019content} proposes novel losses to regularize style encoders to produce disentangled features. To disentangle different attributes, HiSD~\cite{li2021image} decomposes labels into independent tags and employs multiple manipulation networks for each tag separately.
Different from prior arts which disentangle in latent space or generator space, in the setting of designing customizable icons given desired app and theme, our work aims to disentangle two orthogonal concepts w.r.t.\ the theme and app in the discriminator space.

\section{The \dataset Dataset}
\label{sec:dataset}

We propose a novel and high-quality icon dataset with $23,164$ well-organized icons covering $52$ mobile applications and $476$ theme styles, named \dataset. Along with these icon images, we design an orthogonal label system describing the theme and application of each icon. With this label system, highly customizable icon generation models can be developed. Although a few icon datasets have been released, they are mostly not tailored for conditional icon designing tasks. And they do not have enough labels indicating the content and style for each icon, which is vital for icon designing. Table~\ref{tab:dataset comparision} shows difference between \dataset and these datasets.

\begin{table}[t]
    \centering
    \caption{Comparison of different logo/icon datasets.
    `Alpha' refers the transparent channel of images.}
    \label{tab:dataset comparision}
    \vspace{-10pt}
    \begin{tabular}{@{}lcccc@{}}
        \toprule
         Dataset & Images & Labels & Alpha & Source \\
         \midrule
         FlickLogos-32~\cite{romberg2011scalable} & 8k & 32 & - & real world \\
         WebLogo-2M~\cite{su2017weblogo} & 186k & 194 & - & twitter \\
         Logo-2K+~\cite{wang2020logo} & 167k & 2314 & - & real world\\
         LLD-icon~\cite{sage2018logo} & 480k & 0 & - & favicons \\
         LLD-logo~\cite{sage2018logo} & 120k & 0 & - & twitter \\
         icons-50~\cite{hendrycks2018robustness} & 10k & 50 & - & business \\
         Yang \textit{et al}.~\cite{yang2021icon} & 21k & 8 & - & website \\
         \dataset\,(Ours) & 23k & 476$\times$52 &\checkmark & designers \\
         \bottomrule
    \end{tabular}
\end{table}

\begin{figure}[!t]
    \begin{minipage}[b]{.49\linewidth}
      \centering
      \centerline{\includegraphics[width=\linewidth]{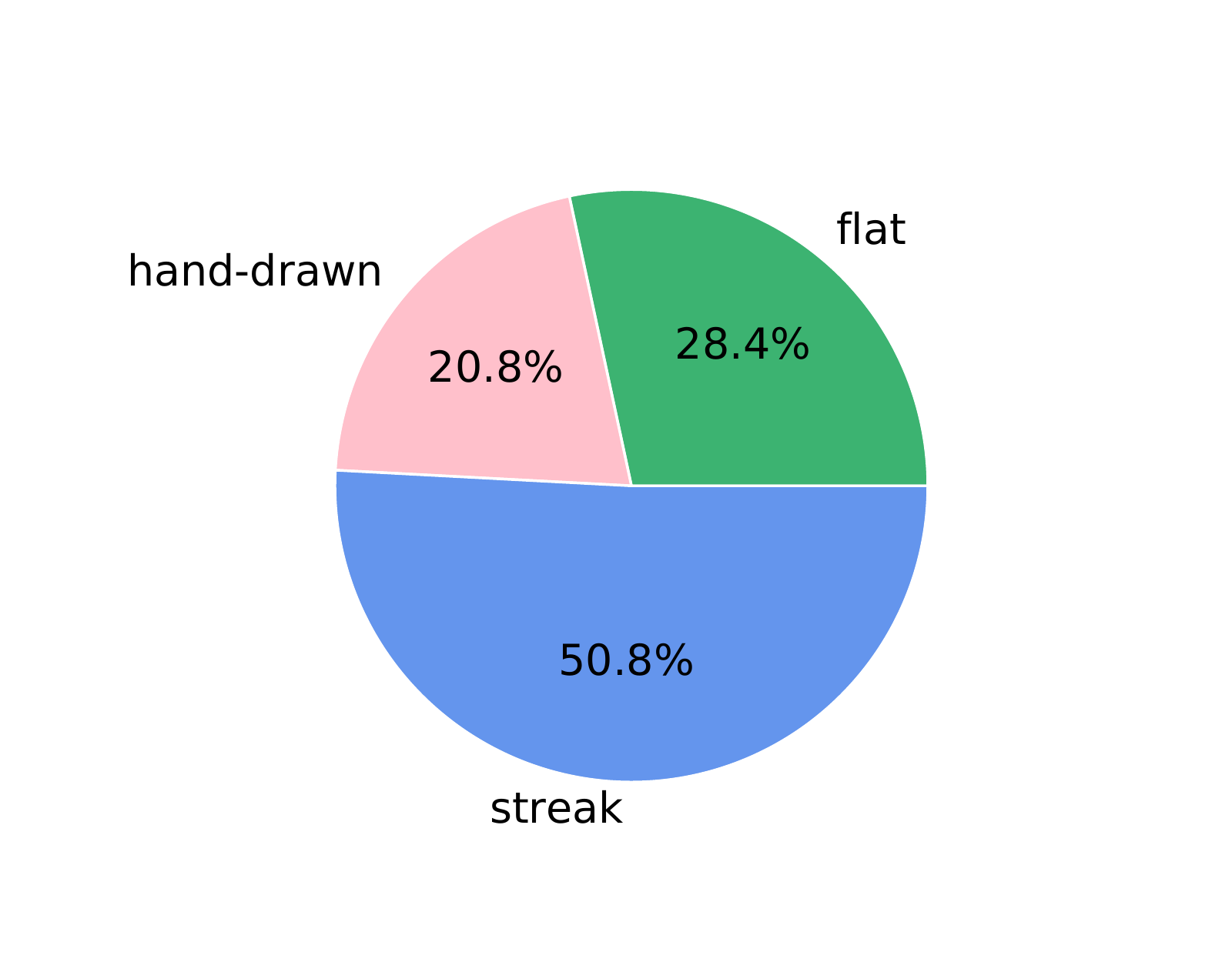}}
      \centerline{\textbf{(a) Overall-styles}}
    \end{minipage}
    \hfill
    \begin{minipage}[b]{.50\linewidth}
      \centering
      \centerline{\includegraphics[width=\linewidth]{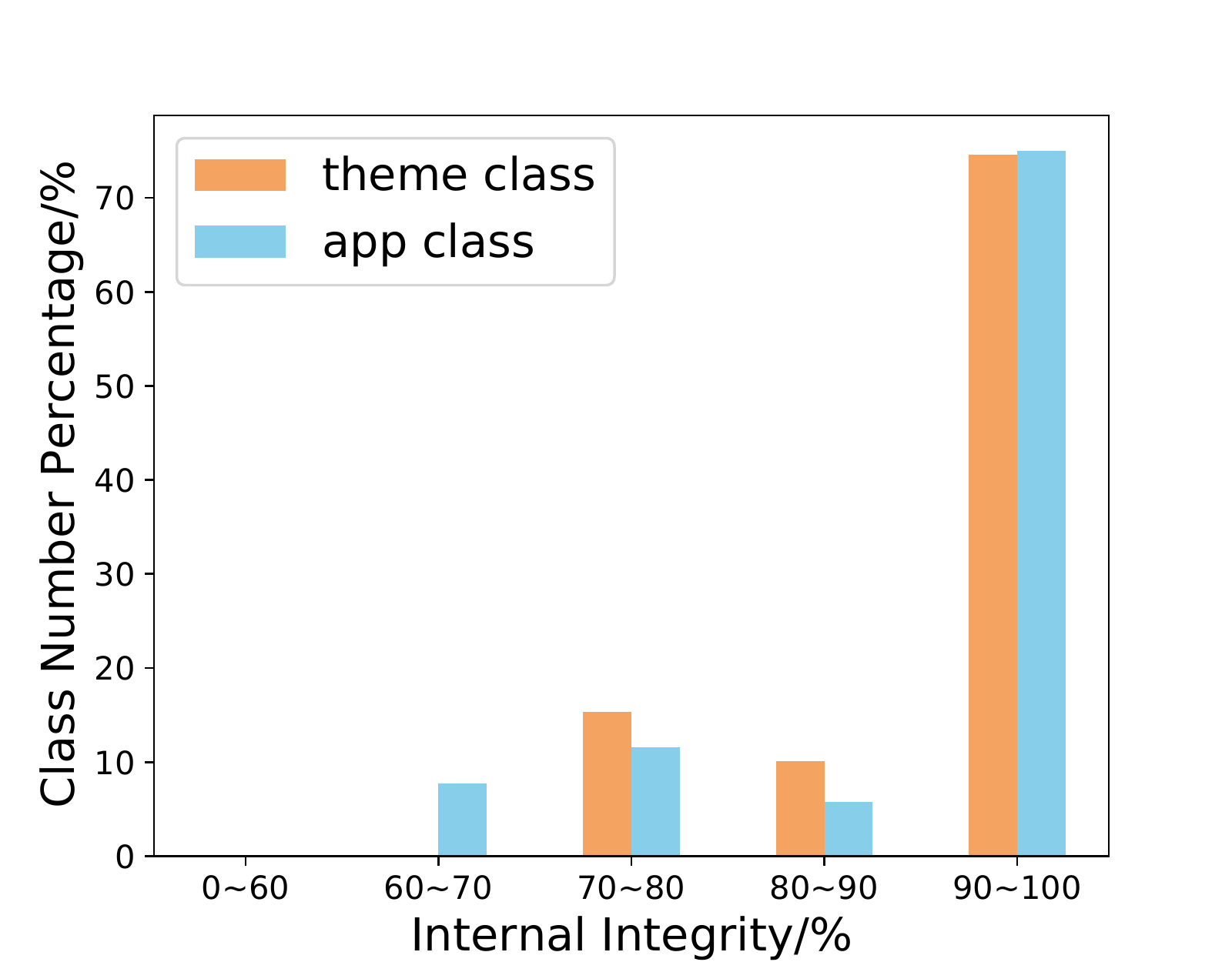}}
      \centerline{\textbf{(b) Integrity check}}
    \end{minipage}
    \vfill
\caption{\dataset dataset. (a) The proportion of three overall-styles. (b) Internal icon integrity of theme and app classes.}
\label{fig:dataset_distribution}
\end{figure}

\vspace{5pt}
\noindent\textbf{Data Collection.} The original image data are collected from different icon designers. We manually check all the samples and discard icons with 1) weird themes, 2) wrong labels, and 3) too many characters. To balance the number of images in each class, apps and themes with few icons are removed.
All the icons are resized from $192\times192$ to $256\times256$ using the Lanczos~\cite{lanczos1950iteration} filter in Pillow. Finally, we yield a clean dataset containing $23,164$ icons. These icons cover $52$ frequently used apps with $476$ different themes. According to different designing approaches, the whole dataset can be divided into three overall-styles: hand-drawn, flat and streak. The proportion of the three overall-styles in the dataset is shown in Fig.~\ref{fig:dataset_distribution}\,(a). Refer to appendix for the visualization of \dataset dataset.

\vspace{5pt}
\noindent\textbf{Orthogonal Label System.} As shown in Fig.~\ref{fig:teaser}\,(a), the proposed \dataset dataset has two orthogonal labels: theme and app label. Here the orthogonality means that the two different labels stand for two completely unrelated concepts: we infer the app of an icon from its contents, such as cloud and sun for weather app, while we infer the theme from the designing styles, e.g., line-styles or colors. Ideally, every individual icon can be retrieved given a pair of theme and app label. But there are always some omissions in the realistic dataset. So we check the integrity of our dataset in Fig.~\ref{fig:dataset_distribution}\,(b). It can be observed $70\%$ of app or theme classes are a full set of icons, and there is no class whose inner icon integrity is below $60\%$. This denotes there is only few absence in our \dataset dataset.

\vspace{5pt}
\noindent\textbf{Differences against natural images.} Compared with commonly-used datasets in generation tasks that consist of realistic photos, images in \dataset dataset are artificially elaborated. Icons from different themes and apps may show completely different appearances. And both the distribution within one class or of overall dataset are not so smooth as that of natural images. These factors render great challenges for the training of GAN models. Besides, due to the orthogonal label system in \dataset dataset, there is only one icon at most given both labels. And this can be viewed as one-shot generation which also challenges conditional generator.

\section{Approach}

\begin{figure}[!t]
    \centering
    \includegraphics[width=0.9\linewidth]{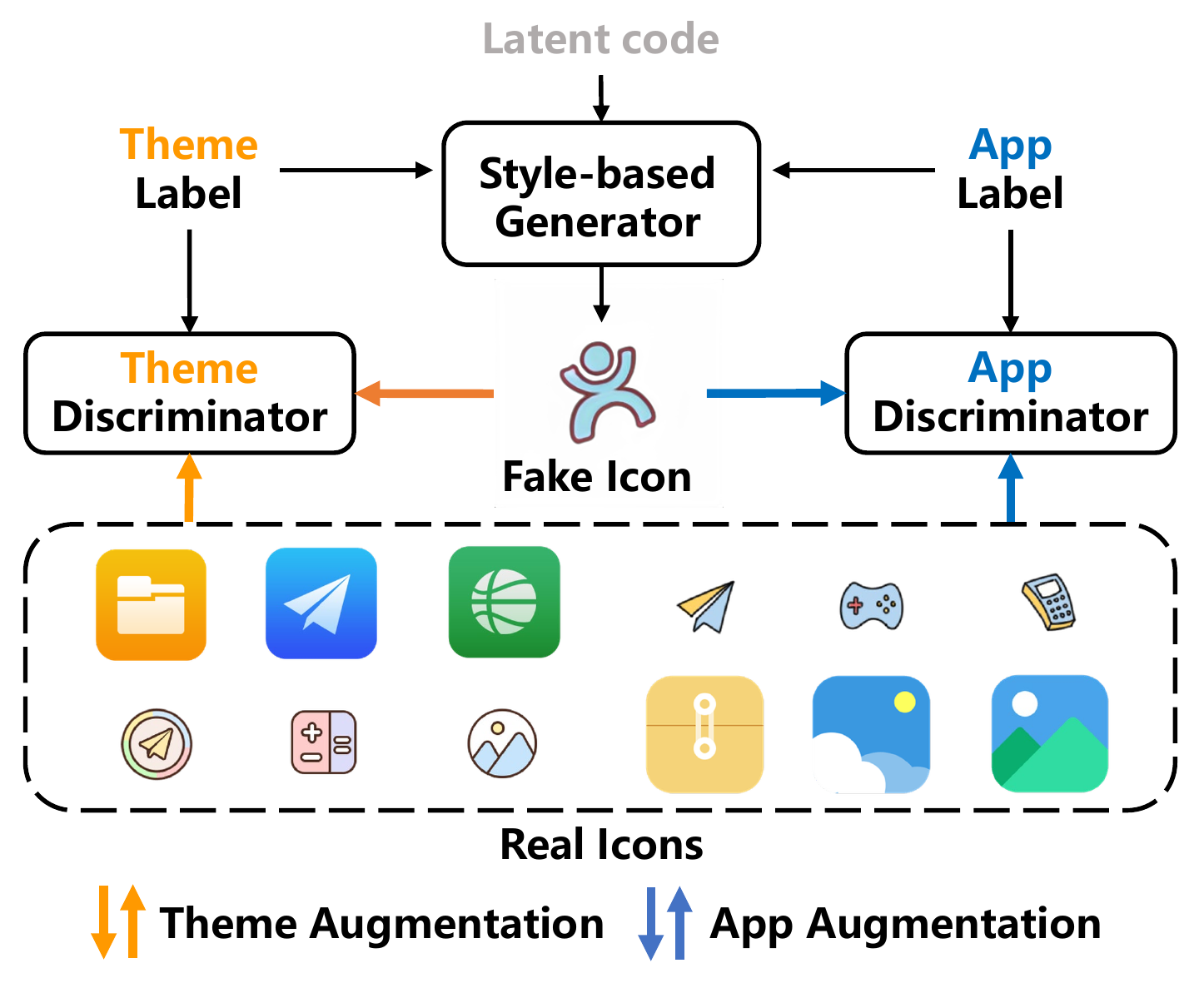}
    \caption{Icon generation with dual discriminators. Each pair of real and generated images are sent to both theme and app discriminators. With irrelevant networks and orthogonal augmentations, the meaning of app and theme labels can be disentangled gradually.}
    \label{fig:method_twobranch}
    \vspace{-5pt}
\end{figure}

\subsection{Motivation Analysis}
\label{sec:motivation}
Our goal is to generate icons that are consistent with desired app and theme labels, so the information of both apps and themes need to be injected into a generative model. Here we first construct a straightforward baseline based on StyleGAN2~\cite{karras2020training}, which comprises a style-based generator $G$ and a discriminator $D$. We simply concatenate one-hot theme and app labels to form a conditional vector $\bm{c}$. For the generator, we send both the conditional vector $\bm{c}$ and random latent code $\bm{z}$ to the mapping network in StyleGAN2. For the discriminator, we follow the projection idea~\cite{miyato2018cgans} by computing the inner product of the embedded condition feature $\bm{v}$ and the intermediate feature $\bm{f}$ as its output. Note that $\bm{v}$ is obtained by processing $\bm{c}$ with a learnable linear projection, and $\bm{f}$ denotes the encoded image feature from the discriminator $D$.

However, as shown in Fig.~\ref{fig:teaser}(b), we find this baseline fails to generate meaningful or high-quality icons. The trained model simply tends to reproduce the original icons without creating new contents or styles. And the quality of generated icons deteriorates significantly when the given app-theme label pair does not exist in the dataset. This indicates this baseline fails to understand theme styles and app contents individually. Instead, it recognizes the concatenation of app and theme label as a joint style condition and learns a malformed relation between the icons and the concatenated labels. Such a problem can be viewed as an entanglement of two orthogonal labels. Even worse, the orthogonality of the label space exacerbates such an entanglement problem since there is only one image conditioned on a concatenated theme and app label. To address this, we propose a dual-discriminator setting and a contrastive feature disentanglement loss. 

\begin{figure}[!t]
    \centering
    \includegraphics[width=0.75\linewidth]{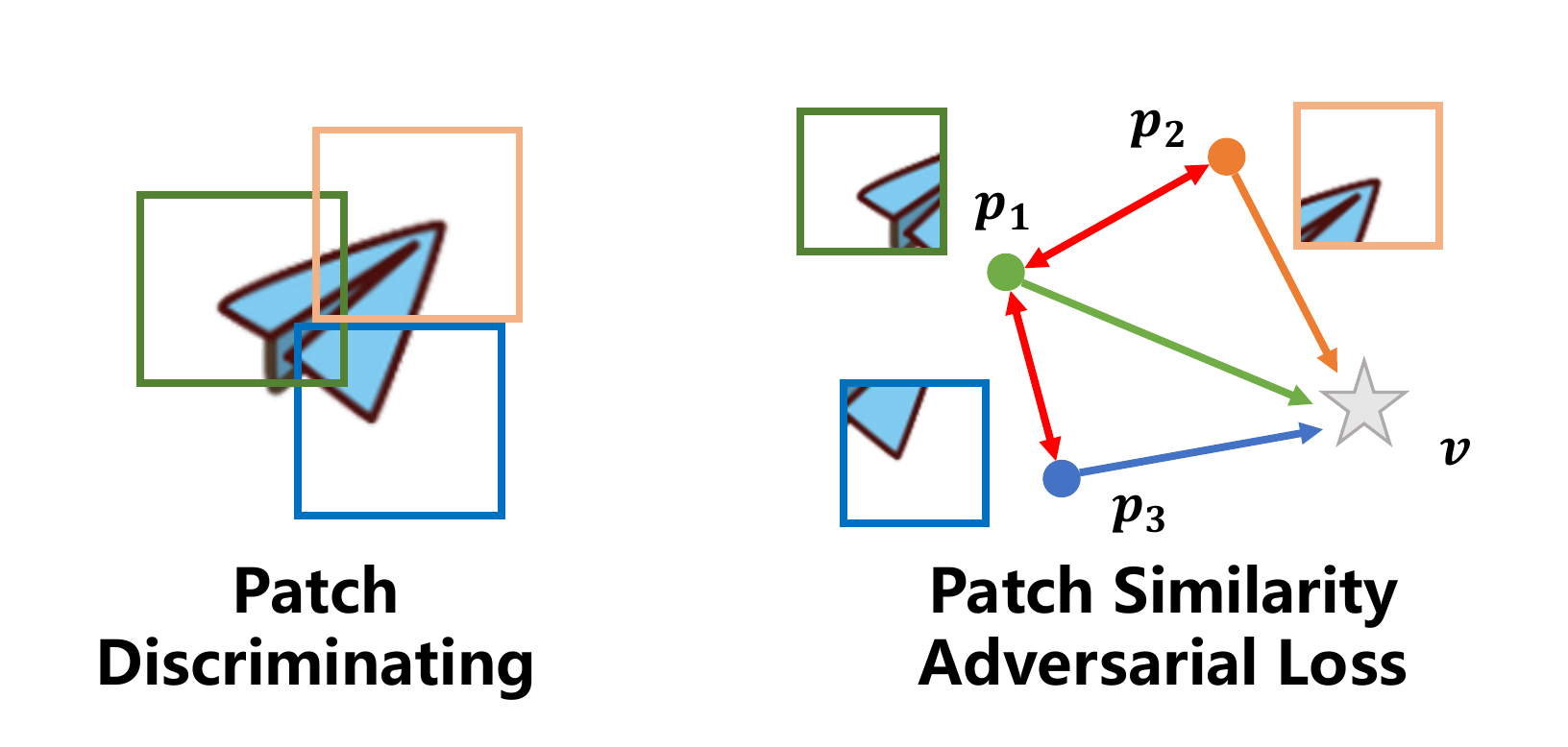}
    \caption{Patch similarity adversarial loss. The theme discriminator extracts different features from image patches to enrich the data from one theme. And the loss applies constraints on patch features from two aspects: to be close to the embedded condition feature and to be similar with each other for correct and spatially-invariant representation.}
    \label{fig:method_patch}
    \vspace{-5pt}
\end{figure}

\subsection{IconGAN with Dual Discriminators}
In the baseline, only one discriminator is adopted to align the generated icons with the real ones. The discriminator learns to mimic the contents of apps and styles of themes simultaneously, which easily induces the entanglement between two concepts. To avoid the entanglement problem caused by network architecture, we propose to use two different discriminators: one for content alignment defined by app labels; the other for style alignment of themes. 

As shown in Fig.~\ref{fig:method_twobranch}, our network consists of a generator $G$, an app discriminator $D_{app}$, and a theme discriminator $D_{thm}$. When training, the generator receives both labels and outputs corresponding icons, and each discriminator concentrates on discriminating just one type of label. In this way, the independent characteristics of app and theme can be better captured. And each discriminator learns one condition representation from a data distribution rather than a single icon, which also helps to reduce the entanglement.

Beyond that, the dual discriminator architecture enables us to apply two types of augmentation for each icon image. Different from existing augmentations used in GANs, which is mainly designed to help overcome overfitting of the discriminator, we design two orthogonal  augmentations to further disentangle two discriminators: i) using x-flip and rotation for the theme discriminator to encourage capturing direction-consistent theme information; ii) using scaling and color transformation for the app discriminator to encourage capturing color-independent and size-independent app information. Note that we apply these two augmentations to both real samples and generated images for non-leaking learning~\cite{karras2020training}. With these additional augmentations, app and theme discriminators can gradually learn more accurate and disentangled representations.

In addition, we employ the patch discriminator~\cite{isola2017image} to $D_{thm}$ for expanding few training samples\,(at most 52) of one theme to more image patches. Thanks to the architecture of separated discriminators, both the global and local information can be supervised by app and theme discriminators respectively and simultaneously. Furthermore, to make the extracted theme information spatially invariant for the theme discriminator and to improve the style consistency in one icon for the generator, we propose a novel patch similarity adversarial loss based on patch discriminator. As shown in Fig.~\ref{fig:method_patch}, different from the 
widely-used PatchGAN~\cite{isola2017image} which only constrains the patch features $\bm{p}$ with embedded condition feature $\bm{v}$ independently, we 
additionally pull together different patch features extracted from one image. Here we define the conditional adversarial loss and patch similarity adversarial loss as 
\begin{equation}
\begin{aligned}
    \mathcal{L}_{adv}=& \underset{x \sim p_d}{\mathbb{E}}\left[\log(D_{app}(x, c_x))\right]\, + \\
    &\underset{z \sim p_z}{\mathbb{E}}\left[\log(1 - D_{app}(G(z, c), c)\right]\,,
\end{aligned}
\end{equation}
\begin{equation}
\begin{aligned}
    \mathcal{L}_{sim-adv}=& \underset{{x \sim p_d}}{\mathbb{E}}\ \underset{{0 \leq i < j \leq M}}{\mathbb{E}}\left[\log(\overline{\bm{p}_i} \cdot \overline{\bm{p}_j} + \bm{p}_i \cdot \bm{v}_{c_x})\right]\, + \\
    &\underset{x \sim p_g}{\mathbb{E}}\ \underset{{0 \leq i < j \leq M}}{\mathbb{E}}\left[\log(1 - (\overline{\bm{p}_i} \cdot \overline{\bm{p}_j} + \bm{p}_i \cdot \bm{v}_{c_x}))\right]\,,
\end{aligned}
\end{equation}
where $M$ is the number of patches, $c$ is the label input, $\overline{\bm{p}}$ represents the $\ell_2$-normalized patch features, and $\bm{v}_{c_x}$ denotes the embedding vector w.r.t.\ the theme class of image $x$. The generator is trained to minimize both adversarial losses, while app and theme discriminators are trained to maximize $\mathcal{L}_{adv}$ and $\mathcal{L}_{sim-adv}$ respectively.

\begin{figure}[!t]
    \centering
    \includegraphics[width=\linewidth]{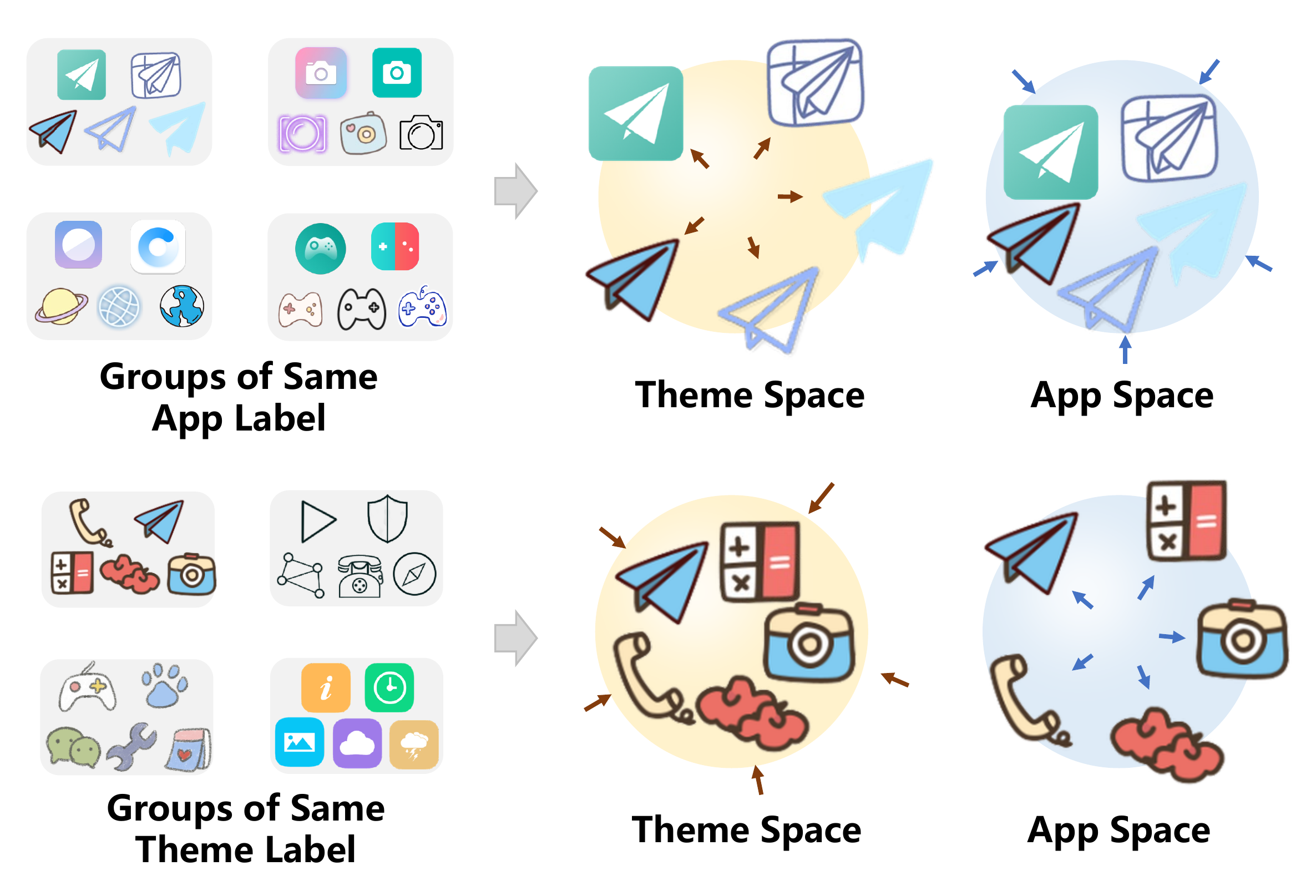}
    \caption{Contrastive feature disentanglement. Given different icon groups sampled from a minibatch, we pull the positive and push the negative discriminator features on two unit hyperspheres within each group.}
    \label{fig:method_contra}
    \vspace{-5pt}
\end{figure}

\subsection{Contrastive Feature Disentanglement}
\label{sec:contrastive feature disentanglement}
Equipped with dual-discriminator architecture, the encoded feature from app and theme discriminators are still cannot be guaranteed to be disentangled, as shown in Fig.~\ref{fig:tsne}. We attribute this problem to the unique distribution characteristics conditioned on app and theme labels in the \dataset dataset. As shown in Fig.~\ref{fig:teaser}\,(a), the icons belonging to one app class vary greatly in different themes, which makes it very hard to capture its core representation by the app discriminator. Conversely, one theme style is easier to learn due to the more pronounced inner distribution patterns, but it is also difficult for the theme discriminator to capture the subtle differences among various similar styles. Even for humans, it is not easy to accurately understand the unique and decoupled characteristics for each category without a large amount of data to compare. So both discriminators may unwittingly learn inaccurate features with uncorrelated information, which causes the generator to produce mismatched icons occasionally.

To help discriminators independently extract correct features of different app and theme icons. We propose the idea of contrastive feature disentanglement\,(CFD), as illustrated in Fig.~\ref{fig:method_contra}. We first sample the icons in one mini-batch into groups with the same app or theme labels. In the two feature spaces from both discriminators, we pull the app features in the group with same app label to be close and push away their theme features. In contrast, the app features in the group with the same theme label are required to be dispersed, while their theme features are encouraged to be close. Note that icons with the same theme\,(app) label must belong to different app\,(theme) classes due to the orthogonality of \dataset dataset, and we only regularize features within each group, which means there is no constraint on features between different groups. Along with training of IconGAN, only features of real icons are used to regularize discriminators. Specifically, we implement the CFD using the alignment loss and uniformity loss following~\cite{wang2020understanding}, which take the form 
\begin{equation}
\mathcal{L}_{align} = \sum_{s=a,h} \underset{0 \leq k \leq R^s}{\mathbb{E}}\ \underset{0 \leq i < j \leq N_k} {\mathbb{E}}\left[\max(||\,\overline{\bm{f}^s_i} - \overline{\bm{f}^s_j}||^2_2 - \epsilon, 0)\right]\,,
\end{equation}
\begin{equation}
\mathcal{L}_{uniform} = \sum_{s=a,h} \underset{0 \leq k \leq R^{\tilde{s}}}{\mathbb{E}}\ \underset{0 \leq i < j \leq N_k} {\mathbb{E}}\left[\mathrm{e}^{-t||\,\overline{\bm{f}^s_i} \,-\, \overline{\bm{f}^s_j}||^2_2}\right]\,,
\end{equation}
where $\overline{\bm{f}}$ is the normalized icon feature from the discriminator, $s$ is an indicator for an app $a$ or a theme $h$, $R$ represents the groups of icons, $N$ denotes the number of images in the corresponding group, $t$ is the temperature of the contrastive loss, and $\epsilon$ controls the nearest distance between two normalized features. Note that we use the center patch feature $\bm{p}$ as the icon feature $\bm{f}$ when training with patch discriminator.
The final loss is defined by
\begin{equation}
    \mathcal{L} = \mathcal{L}_{sim-adv} + \mathcal{L}_{adv} + \lambda_1\mathcal{L}_{align} + \lambda_2\mathcal{L}_{uniform}\,,
    \label{equa:final loss}
\end{equation}
where $\lambda_1$ and $\lambda_2$ are the hyper-parameters controlling the importance of CFD. The pseudo code of training process is provided in the appendix.

\vspace{10pt}
\noindent\textbf{Difference between CFD and contrastive loss in GANs.} Our CFD is inspired by the conditional contrastive loss in~\cite{kang2020contragan} which takes data-to-data relations into consideration. However, different from their proposed data-to-data cross entropy loss~\cite{kang2021rebooting} or 2C loss~\cite{kang2020contragan} designed to improve the performance of cGANs, CFD loss aims to help the two orthogonal discriminators capture accurate class information. Compared with common contrastive losses that leverage as many samples as possible, our CFD only uses negative samples within each group which accounts for only a small part of all negative samples. This is to prevent discriminators from learning entangled representations caused by pushing away the features with both different app and theme labels.

\section{Experiments}
\subsection{Implementation Details}
We build our model upon StyleGAN2-ada~\cite{karras2020training}, and derive the full method step by step. Most training configurations are kept the same as StyleGAN2~\cite{karras2020analyzing}, including the historical average of generator weights~\cite{karras2017progressive}, the lazy regularization of R1 gradient penalty~\cite{mescheder2018training}, the path length constraint~\cite{karras2020analyzing}, the style mixing regularization~\cite{karras2019style}, and the minibatch standard deviation~\cite{karras2017progressive}. For the adversarial loss, we use the non-saturating logistic form~\cite{goodfellow2014generative}. For the uniformity loss, we set $\lambda_1=1$ and $t=2$. For the alignment loss, we set $\lambda_2=1$; $\epsilon=0.25$ and $\epsilon=0.1$ for app group and theme group, respectively. Especially, the weight for R1 regularization is set to $50$ for stable training, and the patch size in $D_{thm}$ is set to $95$. All the parameters are optimized with Adam~\cite{kingma2015adam} optimizer, whose $\beta_1=0$, $\beta_2=0.99$, $\epsilon=10^{-8}$, and learning rate is set to $0.002$. Training ends when both discriminators have seen $10,000$k images. Each training batch includes $64$ icon images with the shape of $256\times256\times4$, which are evenly distributed on $4$ RTX $3090$ GPUs. 

\subsection{Comparing Methods and Metrics}
\label{sec:evaluation}
\textbf{Comparing Methods.}
We compare our IconGAN with these approaches: (1) StyleGAN2~\cite{karras2020analyzing}, which is a strong generative model applied in many natural scenes. We train it under three experimental settings: i) trained using both app and theme labels\,(default). ii) trained only with app labels. iii) trained only with theme labels. (2) StyleGAN2-ada~\cite{karras2020training}, which proposes an adaptive augmentation mechanism for better results when training with limited data. (3) ReACGAN~\cite{kang2021rebooting}, which proposes a novel data-to-data cross-entropy loss to improve image quality of conditional generation. Note that the loss has been verified its harmony with StyleGAN2 architecture in original paper. We employ the loss under two experimental settings: i) StyleGAN2 with one discriminator. ii) StyleGAN2 with two discriminators for a fair comparison with our CFD loss, marked as ReACGAN$\dagger$. (4) LogoSyn~\cite{sage2018logo}, which is the first to generate icon/logo images with synthetic labels. Here we employ their proposed DCGAN-LC architecture for conditional generation and replace the synthetic labels with real labels in our AppIcon dataset.

\vspace{5pt}
\noindent\textbf{Metrics.}
We quantitatively evaluate the generated icons from three aspects: i) accuracy, which represents whether the generated icons conform to the input app and theme labels. ii) Image quality, which reflects how the generated icons resemble the real icons designed artificially. iii) Diversity, indicating the capability of a model producing diverse icons with certain condition and different latent codes. For accuracy, we use two auxiliary ResNet-50~\cite{he2016deep} classifiers pretrained on \dataset dataset to distinguish which app and theme class the generated icons belongs to respectively. And we report the top1 and top5 accuracy using $20k$ randomly generated icons for both theme and app conditions. For image quality, we employ the inception scores\,(IS)~\cite{salimans2016improved} and Fr${\rm \Acute{e}}$chet Inception Distance\,(FID)~\cite{heusel2017gans} metrics, which are widely-used in GANs. Specifically, we compute FID-all between $50k$ generated icons and all dataset icons using Inception-V3~\cite{szegedy2016rethinking} classifier pretrained on ImageNet. And we compute FID-app and FID-sty for each of app and overall-style\,(hand-drawn, streak and flat) classes. Finally, we report their mean FID, termed mFID-app and mFID-sty respectively. For inception score, we compute it among $50k$ generated icons using the same classifier as of FID. To measure diversity, we employ learned perceptual image patch similarity\,(LPIPS) as the metric and compute it between every pair of 10 generated icons with different latent codes under randomly chosen 1000 conditions. Finally we report the mean LPIPS value of three overall -styles, termed mLPIPS. 

\subsection{Comparisons on AppIcon Dataset}
\begin{figure}
    \centering
    \includegraphics[width=0.95\linewidth]{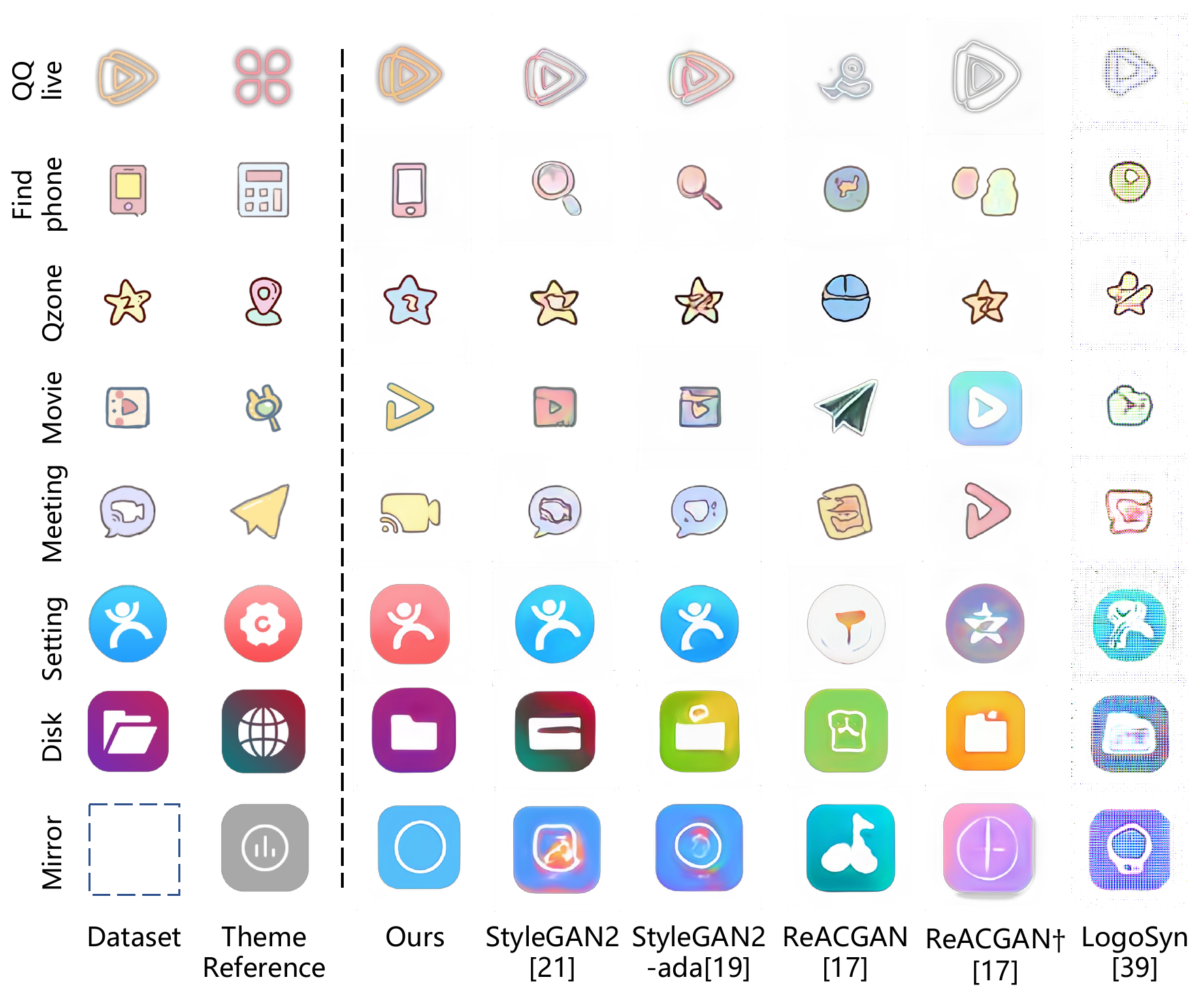}
    \caption{Generated results with random labels. The first column presents the corresponding icons in the dataset. The second column gives a style reference belonging to the same theme class to better illustrate theme styles. Note that the last row shows the generated results of app ``Mirror'' which does not exist in dataset. Zoom in for best visualization.}
    \label{fig:comparison}
    \vspace{-5pt}
\end{figure}

\begin{table*}
  \caption{Quantitative comparison with other methods. The generation of the first two rows use only one label and other rows uses both labels. 
  In each column, the best performance is in bold and the second best is \underline{underlined}.}
  \label{tab:comparison}
  \begin{tabular}{lccccccccc}
    \toprule
     &Top1-thm\,$\uparrow$&Top5-thm\,$\uparrow$&Top1-app\,$\uparrow$&Top5-app\,$\uparrow$&mFID-sty\,$\downarrow$&mFID-app\,$\downarrow$&FID-all\,$\downarrow$&IS\,$\uparrow$&mLPIPS\,$\uparrow$\\
    \midrule
    StyleGAN2\,(app)                   & - & - & \textbf{0.6795} & \textbf{0.8329} & - & 80.23 & 34.45 & 4.48 & - \\
    StyleGAN2\,(thm)                   & 0.1353 & 0.2469 & - & - & 42.39 & - & 32.43 & 4.51 & - \\
    StyleGAN2~\cite{karras2020analyzing}                & 0.1431 & 0.2492 & 0.5964 & 0.7798 & 37.61 & 69.91 & 33.50 & \underline{4.66} & 0.0835\\
    StyleGAN2-ada~\cite{karras2020training}            & \underline{0.1676} & \underline{0.2665} & 0.6010 & 0.7807 & \underline{37.55} & \underline{69.78} & \underline{29.23} & 4.55 & 0.0814\\
    ReACGAN~\cite{kang2021rebooting}            & 0.024  & 0.0721 & 0.0199 & 0.0988 & 48.24 & 97.41 & 36.75 & 4.39 & 0.0903\\
    ReACGAN$\dagger$~\cite{kang2021rebooting} & 0.0193 & 0.0490 & 0.3343 & 0.5992 & 42.87 & 115.52 & 32.15 & 4.32 & \underline{0.1134}\\
    LogoSyn~\cite{sage2018logo} & 0.0130 & 0.0398 & 0.2183 & 0.4264 & 202.86 & 250.72 & 186.52 & 4.14 & 0.0910\\
    Ours                             & \textbf{0.2054} & \textbf{0.3394} & \underline{0.6753} & \underline{0.8290} & \textbf{26.86} &                                            \textbf{61.86}  & \textbf{20.17} & \textbf{4.69} & \textbf{0.1267}\\
    \bottomrule
  \end{tabular}
\end{table*}

\noindent\textbf{Qualitative Comparison.}
Qualitative comparisons of different methods are shown in Fig.~\ref{fig:comparison}. We generate icons with random app and theme labels. The proposed IconGAN can generate correct and clear icons, most of which are quite different from the original ones in the dataset. This indicates IconGAN truly understands different labels and accordingly generate proper contents and styles, even when some conditions do not exist in the dataset. However, StyleGAN2 tends to simply reproduce the original icons without new creations. Even with the adaptive augmentation~\cite{karras2020training}, StyleGAN2-ada shows similar problem. This indicates the entanglement of app and theme labels: the models fail to mimic the theme styles and create app-related contents individually. For the other conditional GAN models, ReACGAN, ReACGAN$\dagger$ and LogoSyn, apparent artifacts in icon images can be observed. Meaningless or wrong contents are generated\,(row 3, 5 and 7), and the styles may be also inconsistent with the required themes\,(row 1 and 4). Comparing with these methods, our IconGAN show superior icon generation capability in terms of both visual harmony, correctness and creativity. 

\vspace{5pt}
\noindent\textbf{Quantitative Comparison.}
Quantitative comparisons are shown in Table~\ref{tab:comparison}. The proposed IconGAN outperforms all the other methods in both image quality and icon diversity. Especially, our method obtains significantly better FID, with a relative improvement of $31\%$ on FID-all comparing with the second best method. From the prospective of generation accuracy, our methods outperforms all the methods which use both app and theme conditions, indicating that our method can generate correct app contents conformed to the desired theme styles. When training uses only one label as condition input for StyleGAN2 models, we find they prefer to generate very similar icons under one specified condition, such as only generating several main apps given one theme label for StyleGAN2\,(thm). Thus they get poor FID and IS scores. Interestingly, StyleGAN2\,(app) get highest app top accuracy but StyleGAN2\,(thm) get a relatively low theme top accuracy. The reason may come from the different distribution of app and theme clustering. Note that ReACGAN gets extremely low generation accuracy as it fails to generate icons corresponding to input labels. The mLPIPS score of ReACGAN$\dagger$ seems good, but it actually yields variously incorrect icons conditioned on certain labels. And the FID performance of LogoSyn is obviously poor because it generates noisy background for icons. We also notice that the IS score is low for all methods. A plausible explanation is that IS metric computation relies on the features extracted by classifiers pretrained on real-world images, which may not be suitable for describing icons. 

\begin{figure}
    \centering
    \includegraphics[width=0.95\linewidth]{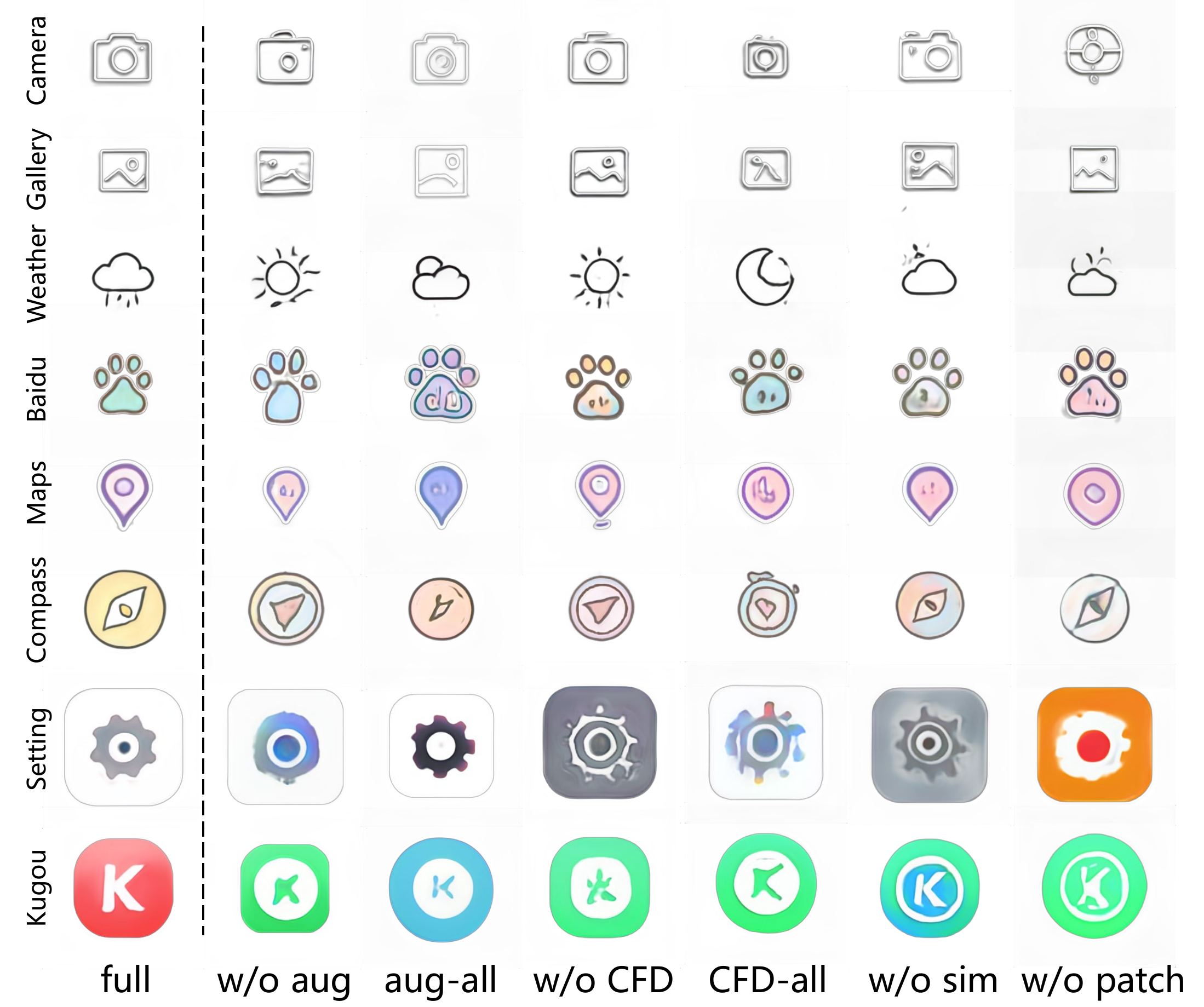}
    \caption{Qualitative results of ablation study. The first column is randomly chosen generated icons with different labels using our IconGAN. The other columns show the results under different configuration.}
    \label{fig:ablation}
    \vspace{-5pt}
\end{figure}

\subsection{Ablation Studies}
Here, we discuss and verify the effectiveness of different components including orthogonal augmentations, CFD loss, patch discriminator and patch similarity adversarial loss. Qualitative and quantitative results are shown in Fig~\ref{fig:ablation} and Table~\ref{tab:ablation}.

\vspace{5pt}
\noindent\textbf{Orthogonal augmentations.} As shown in Table~\ref{tab:ablation}, after removing both augmentations for discriminators\, (w/o aug), generation accuracy drops slightly and the image quality also deteriorates. When applying identically merged augmentation for both discriminators\,(aug-all), the generation accuracy for theme label falls greatly and the other metrics also drops a lot except for mLPIPS, which increases because of the variously mismatched themes in the generated icons. This demonstrates our proposed orthogonal augmentations effectively guides the discriminator focusing on the key representation that can truly distinguish different themes and apps. 

\vspace{5pt}
\noindent\textbf{CFD loss.} Removing CFD loss\,(w/o CFD) from the training loss functions leads to performance drop in all metrics. And the contents of the generated icons become less meaningful, as shown in Fig.~\ref{fig:ablation}\,(row 6 and 8). Then we test a variant loss, CFD-all, where the icons with both different app and theme labels are also used as negative samples to compute the uniformity loss\,(Note CFD loss only uses a small part of negative samples within group of one same label). Such a modification also brings a significant drop in all metrics. These phenomena verify the necessity of our CFD loss which correctly disentangles app and theme information.

\vspace{5pt}
\noindent\textbf{Patch and patch similarity adversarial loss.} When totally removing the patch discriminator and patch similarity adversarial loss\,(w/o patch), from Fig.~\ref{fig:ablation} one can observe that the model fails to learn accurate theme characteristics and cannot produce completely transparent areas in generated icons. Accordingly, theme accuracy and mFID also drop significantly. When only employing patch discriminator without patch similarity adversarial loss\,(w/o sim), we find the color in a generated icon image sometimes not harmonious since there is no consistency constraint among patches of one image, as shown in Fig~\ref{fig:ablation}\,(row 6 and 8). Hence, all metrics deteriorate, especially the theme accuracy. These results illustrate that our patch discriminator is helpful for capturing consistent and accurate style information.
\begin{table*}
  \caption{Quantitative results of ablation study. Best performance is in boldface and second performance is \underline{underlined}.}
  \label{tab:ablation}
  \begin{tabular}{lccccccccc}
    \toprule
    &Top1-thm\,$\uparrow$&Top5-thm\,$\uparrow$&Top1-app\,$\uparrow$&Top5-app\,$\uparrow$&mFID-sty\,$\downarrow$&mFID-app\,$\downarrow$&FID-all\,$\downarrow$&IS\,$\uparrow$&mLPIPS\,$\uparrow$\\
    \midrule
    w/o aug             & \underline{0.1994} & \underline{0.3215} & \underline{0.6694} & 0.8240 & 32.38 & 66.66 & 24.75 & \textbf{4.76} & 0.1138\\
    w/ aug-all          & 0.0207 & 0.0689 & 0.6407 & 0.8022 & 32.21 & 69.34 & \underline{23.01} & 4.46 & \textbf{0.1491}\\
    w/o CFD             & 0.1886 & 0.3040 & 0.6574 & 0.8136 & 32.81 & \underline{64.87} & 25.32 & 4.56 & 0.1049\\
    w/ CFD-all          & 0.1670 & 0.2776 & 0.6530 & 0.8135 & 38.98 & 72.86 & 31.67 & 4.68 & 0.1014\\
    w/o patch           & 0.1647 & 0.2863 & 0.6659 & \underline{0.8248} & \underline{31.20} & 66.76 & 25.36 & 4.59 & 0.1139\\
    w/o sim             & 0.1703 & 0.2931 & 0.6605 & 0.8154 & 33.96 & 66.23 & 25.80 & 4.59 & 0.1082\\
    Ours full           & \textbf{0.2054} & \textbf{0.3394} & \textbf{0.6753} & \textbf{0.8290} & \textbf{26.86} & \textbf{61.86} & \textbf{20.17} & \underline{4.69} & \underline{0.1267}\\
    \bottomrule
  \end{tabular}
\end{table*}
\begin{figure*}
    \centering
    \includegraphics[width=\linewidth]{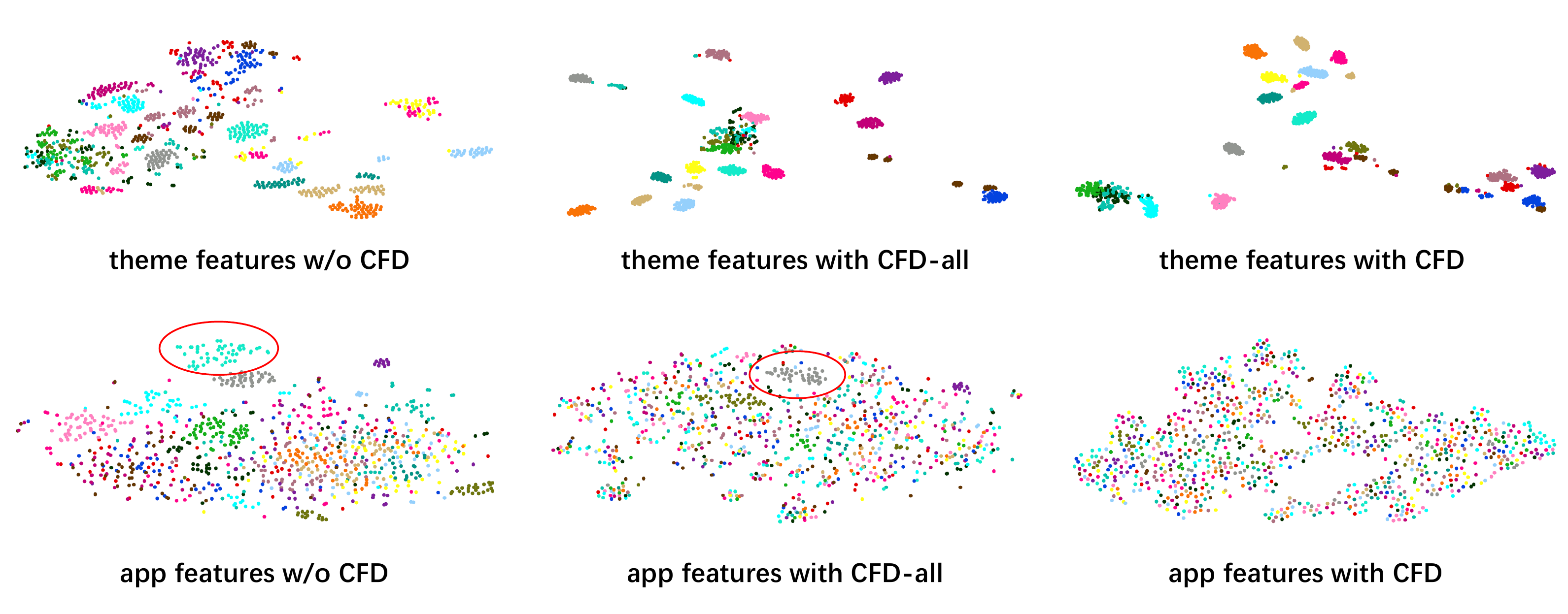}
    \caption{Visualized distribution of discriminator features using t-SNE. The upper plots feature embeddings of theme discriminator. The lower is feature embeddings from app discriminator. Note that different colors represent different themes. Red circles highlight the theme entanglement in app feature space. Best viewed in color.}
    \label{fig:tsne}
\end{figure*}
\begin{figure}
    \centering
    \includegraphics[width=0.95\linewidth]{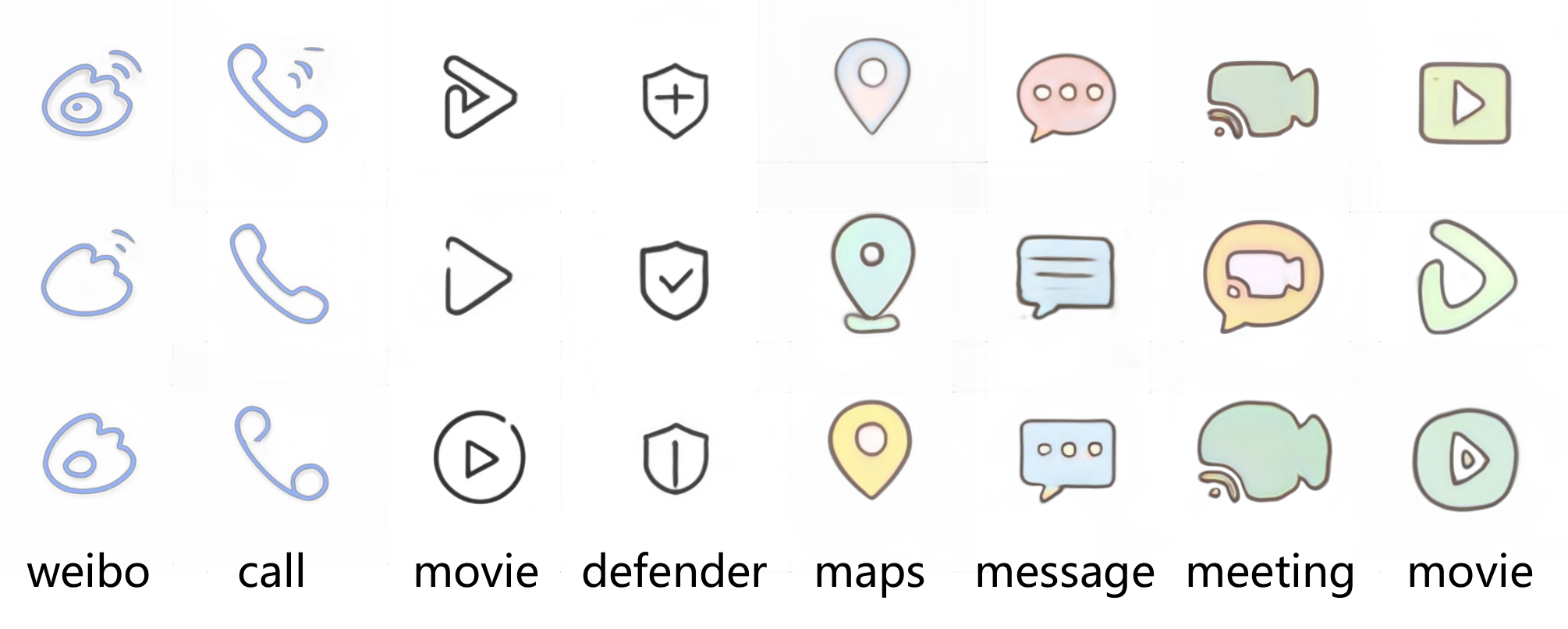}
    \caption{Qualitative results of generation diversity. Each column contains the generated icons with same label and different latent codes using our IconGAN.}
    \label{fig:diversity}
    \vspace{-5pt}
\end{figure}

\subsection{Analysis of Disentanglement.}
\label{sec:analysis}
\noindent\textbf{Feature Level}.
To verify whether the features are truly disentangled, we visualize the distributions of app and theme features from two discriminators via t-SNE~\cite{van2008visualizing} in Fig.~\ref{fig:tsne}. Without CFD, different app features with the same theme labels cluster into groups. This indicates that the learned app features are correlated with the theme information, \textit{i.e.,} the entanglement of app and theme features. After applying CFD, obviously, the app features show no cluster indication with theme labels, demonstrating the success of disentanglement. Besides, the features of different themes become more distinguishable comparing with no CFD. This helps model generate distinct styles instead of mixed styles from various themes, which are common artifacts for other methods. Replacing CFD with CFD-all seems to yield better theme feature clusters, but the entanglement on app features re-appears, which we think is caused by pushing negative samples with both different apps and themes.

\vspace{5pt}
\noindent\textbf{Image Level}.
A good designer can give various designs that satisfy the requirements, so does our IconGAN. Human can represent one app with different contents, \textit{e.g.}, a shopping bag or a stall for ``app store''. If app and theme labels are properly disentangled, the generation results should vary by switching those contents while leaving the style unchanged. We show some generated examples in Fig.~\ref{fig:diversity}, where icons in the same column are with the same conditional labels but different latent codes. The results show that our method can generate icons with different patterns such as various colors or shapes, which also demonstrates that the proposed IconGAN indeed disentangles two orthogonal labels and understands the meaning of app and theme correctly.

\section{Conclusion}
In this work, we introduce a novel icon generation task where icons are required to be generated conditioned on both app and theme labels. To address this new task, we propose a high-quality and well-organized icon dataset annotated with orthogonal theme and app labels. We also 
examine a baseline on the proposed dataset using StyleGAN2 and observe that it fails to produce meaningful and high-quality icons due to the entanglement of app contents and theme styles. To solve the new challenge with orthogonal labels, we propose IconGAN consisting of several novel disentanglement strategies including dual-discriminator architecture and contrastive feature disentanglement, which guides both the generator and discriminator to learn app content creation and theme style alignment separately. Extensive experiments and visualizations demonstrate the superiority of IconGAN and the effectiveness of our disentanglement strategies.

\begin{acks}
This work was funded by the DigiX Joint Innovation Center of Huawei-HUST.
\end{acks}

\bibliographystyle{ACM-Reference-Format}
\bibliography{main.bbl}

\appendix
\vspace{65pt}
\definecolor{CommentGreen}{RGB}{50,100,40}

\section{Training algorithm}

\begin{algorithm}[h]
  \renewcommand{\algorithmicrequire}{\textbf{Input:}}
  \renewcommand{\algorithmicensure}{\textbf{Output:}}
  \renewcommand{\Comment}[1]{\hfill\textcolor{CommentGreen}{$\triangleright$}\textcolor{CommentGreen}{#1}}
  \caption{Training IconGAN.}
  \begin{algorithmic}[1]
  
    \Require 
    Batchsize: N;
    Parameters of $G$: $\theta$; 
    Parameters of $D_{thm}$: $\phi_t$; 
    Parameters of $D_{app}$: $\phi_a$; 
    Feature extractor of $D_{app}$: $Df_{app}$; 
    Feature extractor of $D_{thm}$: $Df_{thm}$; 
    Loss hyperparameters: $\lambda_1, \lambda_2$; 
    Adam hyperparameters: $\beta_1, \beta_2$; 
    Learning rate: $\alpha$.
    
    \Ensure 
    Optimized ($\theta, \phi_a, \phi_t$)
    
    \vspace{10pt}
    \State Initialize $\theta, \phi_a, \phi_t$
    \Repeat
    \State Sample $\bm{z} \gets \{\bm{z}_i\}_{i=1}^N \sim \mathcal{N}(0,1)$
    \State Sample $\bm{\tilde{c}} \gets \{c_i\}_{i=1}^N \sim p_{real}(c)$
    \State Sample $\bm{x}, \bm{c} \gets \{x_i\}_{i=1}^N, \{c_i\}_{i=1}^N \sim p_{real}(x,c)$
    \State $\tilde{\bm{x}} \gets G(\bm{z}, \bm{\tilde{c}})$ \Comment{generate fake images}
    \State Split $\bm{c}$ = $\bm{c}^{a}, \bm{c}^{t}; \bm{\tilde{c}} = \bm{\tilde{c}}^{a}, \bm{\tilde{c}}^{t}$ \Comment{conditions of app \& theme}
    \For{$\{i=1,2...,N\}$} \Comment{sample features to groups}
    \State $align\_apps[\bm{c}_i^a] \gets Df_{app}(\bm{x}_i)$
    \State $unif\_apps[\bm{c}_i^t] \gets Df_{app}(\bm{x}_i)$
    \State $align\_thms[\bm{c}_i^t] \gets Df_{thm}(\bm{x}_i)[center]$
    \State $unif\_thms[\bm{c}_i^a] \gets Df_{thm}(\bm{x}_i)[center]$
    \EndFor
    \State $\mathcal{L}_{adv} \gets \mathcal{L}_{adv}(\bm{c^a}, \bm{\tilde{c}^a}, \bm{x}, \tilde{\bm{x}})$
    \State $\mathcal{L}_{sim-adv} \gets \mathcal{L}_{sim-adv}(\bm{c^t}, \bm{\tilde{c}^t}, \bm{x}, \tilde{\bm{x}})$
    \State $\mathcal{L}_{align} \gets \mathcal{L}_{align}(align\_thms, align\_apps)$
    \State $\mathcal{L}_{unif} \gets \mathcal{L}_{unif}(unif\_thms, unif\_apps)$
    \State $\mathcal{L}_{CFD} = \lambda_1\mathcal{L}_{unif}+\lambda_2\mathcal{L}_{align}$
    \State $\theta \gets Adam(\mathcal{L}_{adv}+\mathcal{L}_{sim-adv}, \alpha, \beta_1, \beta_2)$
    \State $\phi_a \gets Adam(\mathcal{L}_{adv}+\mathcal{L}_{CFD}, \alpha, \beta_1, \beta_2)$
    \State $\phi_t \gets Adam(\mathcal{L}_{sim-adv}+\mathcal{L}_{CFD}, \alpha, \beta_1, \beta_2)$
    \Until{Discriminator has seen $10,000k$ real images.}
    
  \end{algorithmic}
\end{algorithm}

\section{Overall Architecture}
\begin{figure}
    \centering
    \includegraphics[width=\linewidth]{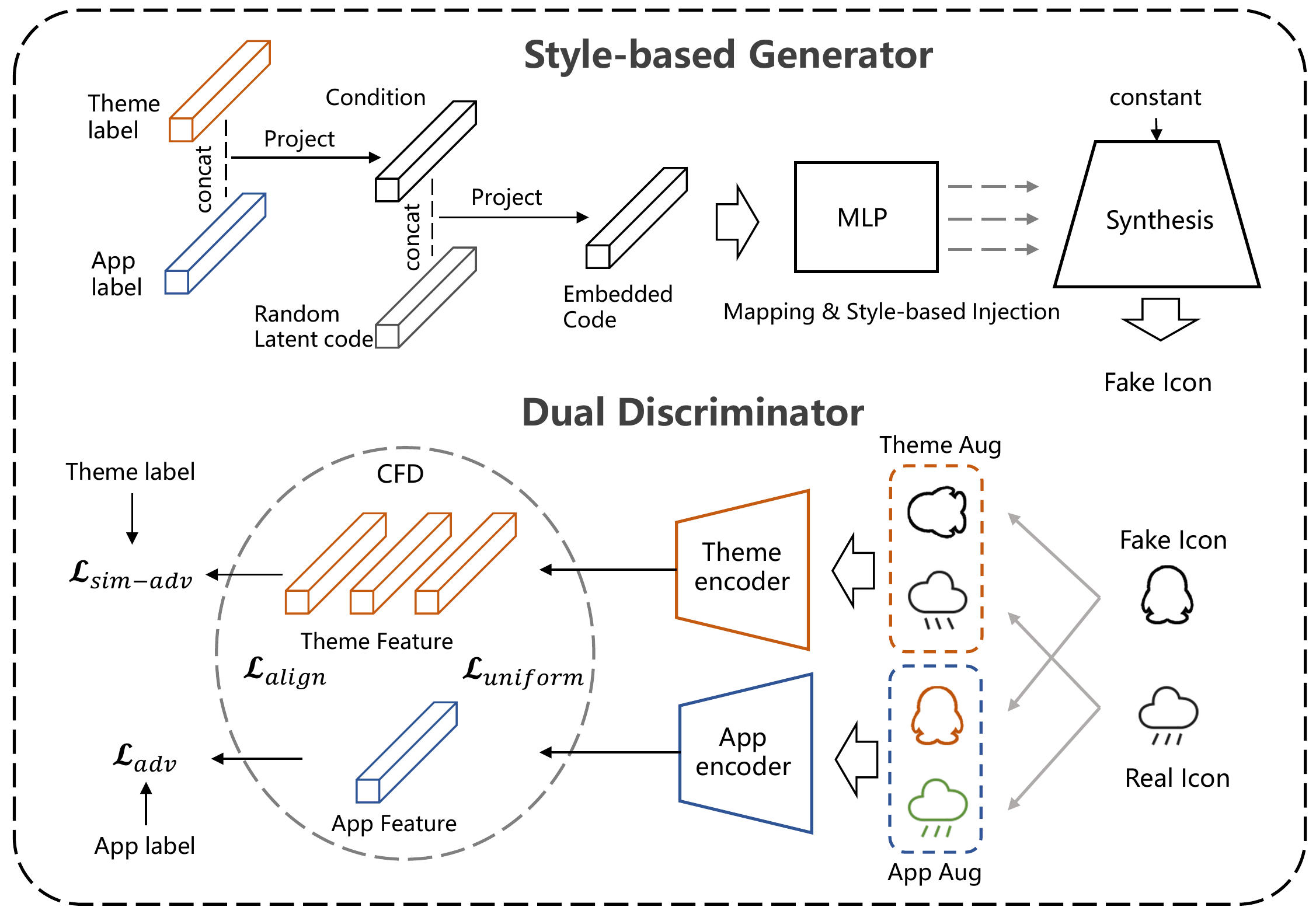}
    \caption{The Overall Architecture of IconGAN. The orange color represents theme-related features or networks and the blue color represents app-related ones. Gray dotted circles indicates where the contrastive feature disentanglement constrains. Zoom in for best visualization.}
    \label{fig:pipline}
\end{figure}

Fig.~\ref{fig:pipline} presents the overall architecture of our IconGAN. The generator is a conditional style-based generator: Input labels(one-hot app and theme labels) are first concatenated and then projected by a linear mapping layer to obtain the condition vector $\textbf{c}$; The embedded code which is obtained by combining random latent code with condition vector is sent to the mapping network to produce style codes; Finally the style codes are injected into the synthesis network to generate icon images which is the same way as StyleGAN2~[21]. And the dual discriminators are two separate models: After passing orthogonal augmentations, fake and real icon images are sent to dual-discriminators to produce corresponding features for computing adversarial and contrastive losses. The encoders of two discriminators are composed of several residual blocks with leaky relu activations.

\section{Dataset Visualization}
\begin{figure*}[t]
    \centering
    \includegraphics[width=\linewidth]{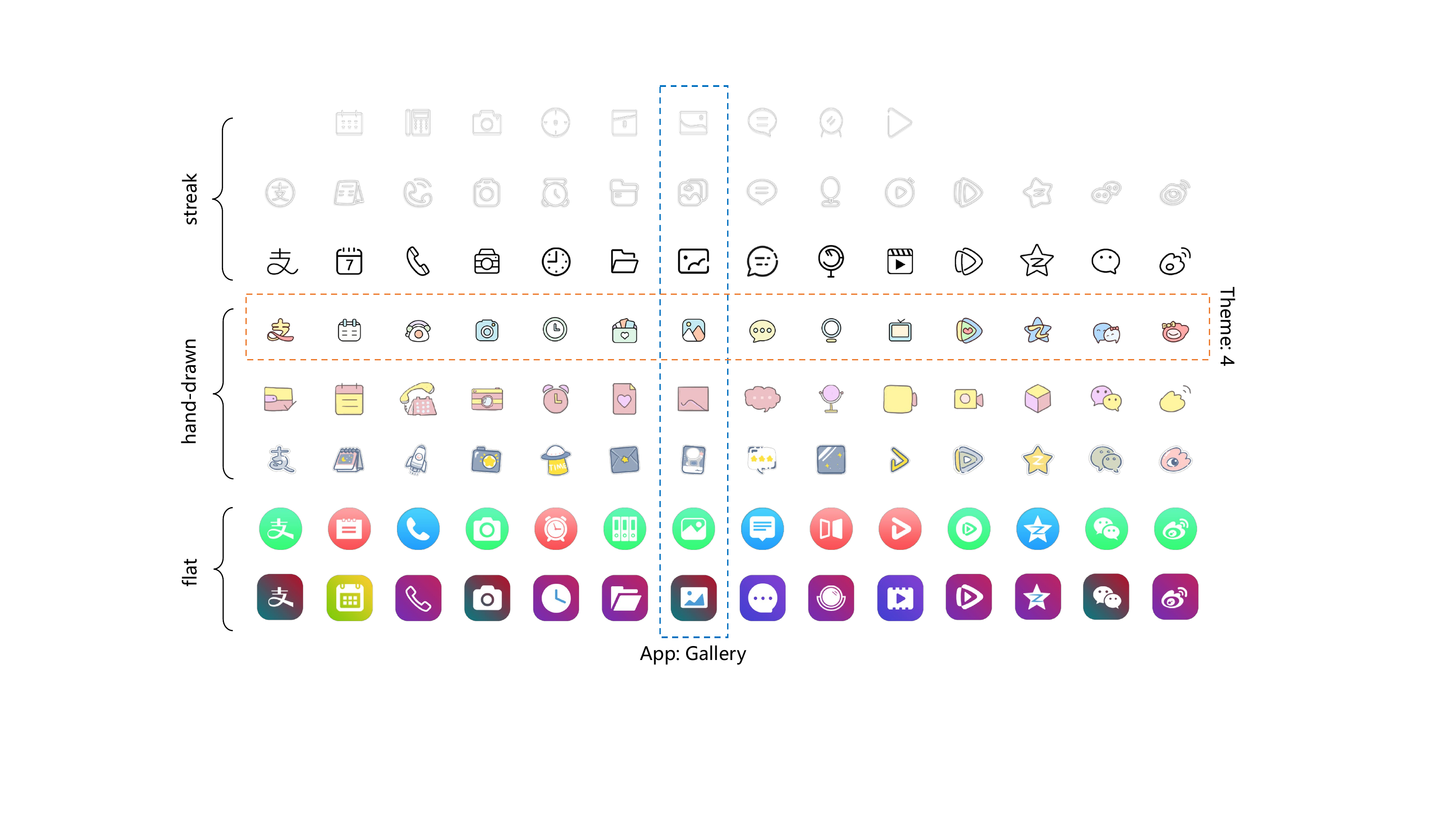}
    \vspace{-5pt}
    \caption{\textbf{Visualization of AppIcon dataset.} The icons in each column are the same mobile application, and the icons in each row share same theme style. The dotted orange and blue boxes highlight an example. The rows from top to bottom correspond to three different overall-styles in order. Note that the blanks denote there is no such an icon for that app-theme combination. }
    \label{fig:supple_dataset}
\end{figure*}

As shown in Fig.~\ref{fig:supple_dataset}, we manually divide the whole dataset to three overall-styles: i) streak overall-style, where the icons are mostly made of geometric lines in white or black colors. ii) hand-drawn overall-style, where the icons are mainly composed of hand-drawn edges filled with several colors. iii) flat overall-style, where the icons are square with a monotone background. Due to the different distribution of three overall-styles, we compute FID and LPIPS metrics on each overall-style individually and calculate the mean value of them as the final mFID-sty and mLPIPS.

\section{User Study}
To more comprehensively assess generation results of different models, we conduct an additional user study. In detail, we randomly choose $20$ app-theme combinations and one fixed noise latent code as the input of all models, thus generating $100$ icon images. $52$ participants are invited to score each icon image by making one choice from five options("excellent", "good", "not bad", "poor" and "meaningless") according to the generation accuracy and quality. As Table~\ref{tab:userstudy} shows, user satisfaction of our results is obviously higher than the others, which demonstrates the superiority of our method on a subjective level.

\begin{table}[hb]
    \centering
    \caption{User Study Results. Satisfied and unsatisfied represent the proportion of the answers above "good" and below "poor" respectively. And the score is calculated by averaging all answers with different weights\,(0\textasciitilde10).}
    \label{tab:userstudy}
    \begin{tabular}{lccc}
        \toprule
         & Satisfied & Unsatisfied & Score\\
        \midrule
         StyleGAN2~[21] & 35.63\% & 2.38\% & 5.62 \\
         StyleGAN2-ada~[19] & 33.13\% & 6.50\% & 5.29 \\
         ReACGAN~[17] & 30.88\% & 5.75\% & 5.07 \\
         LogoSyn~[39] & 12.12\% & 42.13\% & 2.38 \\
         Ours    & \textbf{64.00\%} & \textbf{0.75}\% & \textbf{7.48} \\
        \bottomrule
    \end{tabular}
\end{table}

\begin{figure*}[t]
    \centering
    \includegraphics[width=0.9\linewidth]{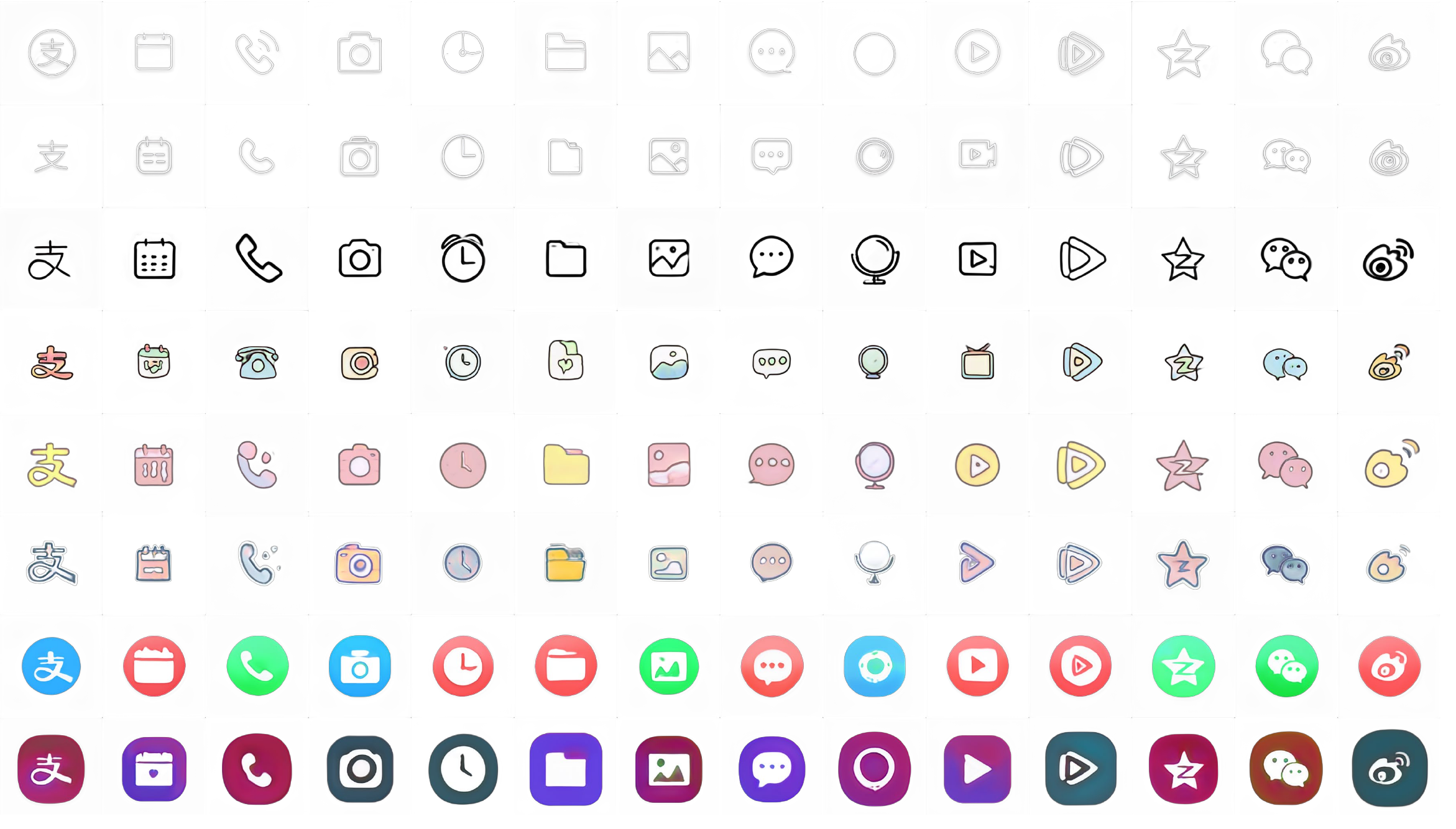}
    \vspace{-5pt}
    \caption{Generated results by IconGAN, corresponding to the dataset icons displayed in Fig.~\ref{fig:supple_dataset}.}
    \label{fig:supple_results}
\end{figure*}

\end{document}